\documentclass[runningheads]{llncs}

\usepackage{eccv}

\usepackage{eccvabbrv}

\usepackage{graphicx}
\usepackage{booktabs}

\usepackage[accsupp]{axessibility}  

\usepackage{xcolor}
\usepackage[colorlinks=true, citecolor=blue!20!cyan]{hyperref}

\hypersetup{
  colorlinks=true,
  linkcolor=red,
  citecolor=blue!20!cyan,
}

\usepackage{orcidlink}
\usepackage{multirow}
\usepackage{booktabs}
\usepackage{colortbl}
\usepackage{amssymb}
\usepackage{ulem}
\usepackage{mathtools}
\usepackage{subcaption}
\usepackage{ulem}
\usepackage{algorithm}
\usepackage{algpseudocode}
\usepackage{newtxmath}
\usepackage{float}
\begin{document}

\title{FALIP: Visual Prompt as Foveal Attention Boosts CLIP Zero-Shot Performance} 

\titlerunning{Foveal Attention CLIP}

\author{Jiedong Zhuang \and
Jiaqi Hu \and
Lianrui Mu \and
Rui Hu \and
Xiaoyu Liang \and
Jiangnan Ye \and
Haoji Hu}

\authorrunning{Zhuang et al.}

\institute{Zhejiang University \\
\email{\{zhuangjiedong,haoji\_hu\}@zju.edu.cn}}

\maketitle

\begin{abstract}
CLIP has achieved impressive zero-shot performance after pretraining on a large-scale dataset consisting of paired image-text data. Previous works have utilized CLIP by incorporating manually designed visual prompts like colored circles and blur masks into the images to guide the model's attention, showing enhanced zero-shot performance in downstream tasks. Although these methods have achieved promising results, they inevitably alter the original information of the images, which can lead to failure in specific tasks. We propose a train-free method \textbf{F}oveal-\textbf{A}ttention C\textbf{LIP} (\textbf{FALIP}), which adjusts the CLIP's attention by inserting foveal attention masks into the multi-head self-attention module. We demonstrate FALIP effectively boosts CLIP zero-shot performance in tasks such as referring expressions comprehension, image classification, and 3D point cloud recognition. Experimental results further show that FALIP outperforms existing methods on most metrics and can augment current methods to enhance their performance. Our project page is link to \href{https://pumpkin805.github.io/FALIP/}{https://pumpkin805.github.io/FALIP/}.
\end{abstract}

\section{Introduction}

Vision-Language Models (VLMs) like CLIP\cite{CLIP} have shown remarkable zero-shot performance in various tasks without further training\cite{exploiting, pointclip,zerocap,cris,maskclip}.
To further expand CLIP's capability, researchers have explored strategies to manually craft input prompts to CLIP. While previous works mainly focus on {\it text} prompts inspired by research in large language models (LLMs), {\it visual} prompts have been recently introduced\cite{redcircle,CPT,FGVP,reclip,SAM}, utilizing symbols such as boxes, points, circles, masks, and others to give models additional cues. These techniques achieve various levels of success on tasks including referring expression comprehension\cite{redcircle,CPT,FGVP,reclip}, part detection\cite{FGVP} and keypoint matching\cite{redcircle}.

Despite the promising results on several tasks, we lack a systematic and intuitive understanding of why visual prompts are effective in improving the zero-shot capability of CLIP. To investigate the mechanisms behind the success of manually designed visual prompts, we conduct an in-depth exploration, starting from analyzing the effectiveness of visual prompts on CLIP over the task of referring expressions comprehension\cite{refcoco,refcocog}. 
Our objectives are twofold: 1) a more principled understanding of the effectiveness of visual prompts, and 2) to design more effective strategies to enhance the zero-shot capability of CLIP.

\begin{figure}[t]
\centering
\includegraphics[scale=0.85]{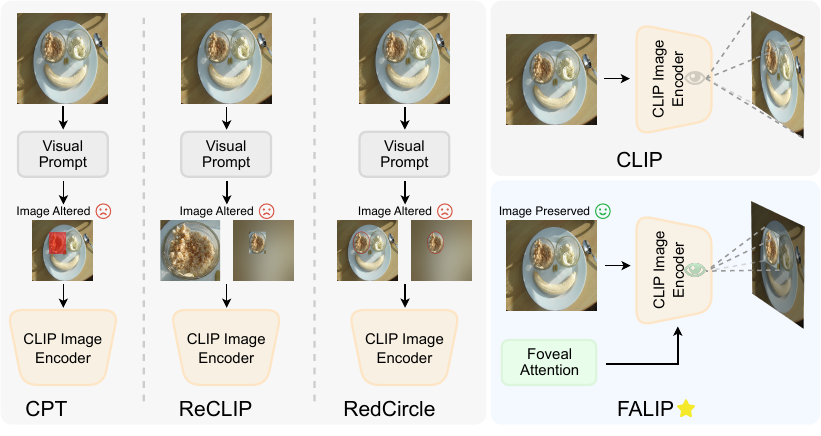}
\put(-310,4.5){\cite{CPT}}
\put(-235,4.5){\cite{reclip}}
\put(-159,4.5){\cite{redcircle}}
\put(-66,106.5){\cite{CLIP}}
\caption{Overview of visual prompt based methods and FALIP. \textit{\textbf{Left}} is the the visual prompt methods\cite{CPT,reclip,redcircle}. They perform image editing (such as colored boxes, cropping, circles, blur masks, etc.) enabling CLIP to perceive specific regions. \textbf{\textit{Bottom right}} is FALIP. It does not alter the content of the original image. The gray dashed line represents the attention of model. Compared to the original CLIP, FALIP aligns more with human visual characteristics. 
}
\label{fig:0}
\end{figure}
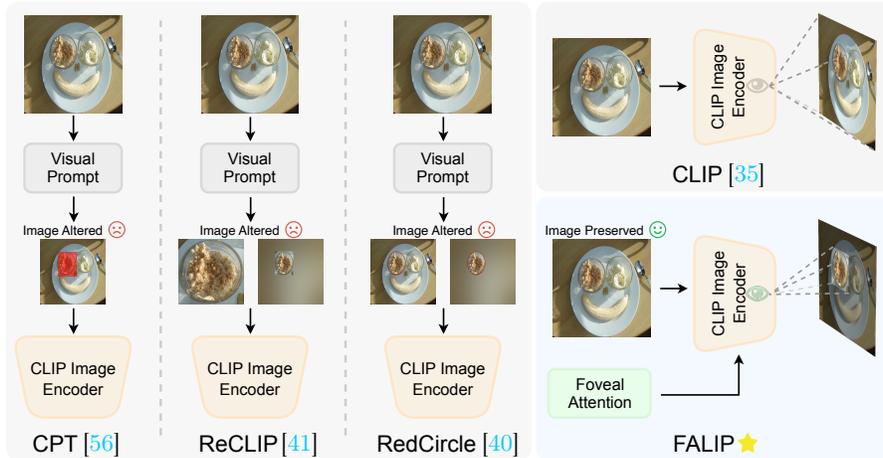

\begin{figure}[t]
\centering
\includegraphics[scale=1.1]{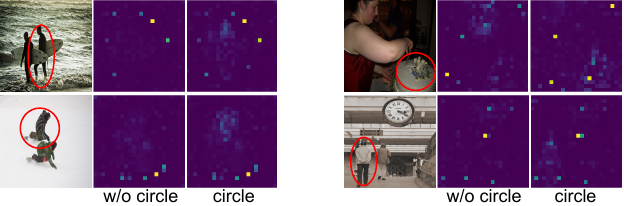}
\caption{The shift in the model's attention before and after incorporating visual prompts. It can be observed that visual prompts can guide the model's attention to specific regions.
}
\label{fig:2}
\end{figure}

In our study, we first examine CLIP's attention maps on numerous images. \cref{fig:2} notes a clear link between visual prompts and model focus: \textit{model attention often zeroes in on areas marked by the visual prompt}.
Using a state-of-the-art visual prompt technique (RedCircle\cite{redcircle}) for zero-shot classification with CLIP, we unexpectedly find that its zero-shot efficacy diminishes with the visual prompt in place. This outcome prompts us to re-think the task-specific effectiveness of such visual prompt methods.
Crucially, these methods edit the image directly, potentially compromising its integrity by occluding or destroying vital details in the image.
For example, RedCircle introduces additional red elements, which could potentially skew fine-grained classification outcomes. Likewise, the ``blur mask'' obfuscates much of the image, retaining only basic shapes and thus discarding significant detail in certain areas. Consequently, these approaches may be ineffective in scenarios demanding high image fidelity.
Our discoveries highlight a paradox in current visual prompt strategies: although aiming to direct CLIP's focus to particular image areas, they inadvertently strip away crucial content, undermining the model's performance. This raises an essential question: is it possible to leverage visual prompts' advantages without sacrificing the integrity of the input image?

In this paper, we introduce \textbf{F}oveal-\textbf{A}ttention C\textbf{LIP}(\textbf{FALIP}), a novel approach that aligns regions of attention (ROA) in images with their corresponding token positions, constructing a foveal attention mechanism into the model's self-attention layer.
Drawing inspiration from human visual perception \cite{unsupervised-foveal,fovea-drive} featuring selective focus and specific region processing, FALIP enhances CLIP with similar attentional characteristics. \cref{fig:0} provides a concise illustration comparing the attention mechanisms of CLIP and human cognition, as well as a comparison between our method and existing techniques.
FALIP has been rigorously tested across numerous datasets, demonstrating competitive performance. Remarkably, it is designed to be plug-and-play, adding negligible computational cost, requiring no further training, and complementing existing approaches. 
In addition, through our experiments, we uncover that the CLIP model attention heads vary in their response to visual prompts, and we find that adjusting these heads may further unleash the effectiveness of visual prompts.

In summary, our main contributions can be outlined as follows: 
(1) We propose FALIP, a novel method to adaptively guide the attention of CLIP during inference without additional training. 
(2) We extensively evaluate FALIP on a wide range of tasks and datasets and achieve competitive performance compared to existing methods.
(3) We present an in-depth analysis that demystifies the surprising effectiveness of visual prompts and sheds new light on improving the zero-shot inference capability of CLIP.
(4) We discover that different attention heads in the CLIP model exhibit varying levels of sensitivity to visual prompts, and they can be adjusted to unleash the full potential of visual prompts.

\section{Related Work}

\textbf{Vision-Language Models.} 
CLIP\cite{CLIP} uses massive amounts of image-text paired data for contrastive learning, enabling it to acquire powerful zero-shot image classification and image-text retrieval capabilities. Its introduction also further propelled the subsequent development of models such as Multimodal Large Language Models (MLLMs), BLIP\cite{BLIP,BLIP2}, and LLaVA\cite{LLava,LLava2,llama,llama2} that built upon the foundation established by CLIP. Although recent other vision models based on ViT\cite{ViT}, such as DINO\cite{dino,dinov2}, MAE\cite{mae}, and MoCo\cite{mocov3}, have achieved remarkable performance in single-modal visual tasks, they do not possess the cross-modal capabilities of CLIP. This paper focuses on CLIP and proposes a method that can further enhance the zero-shot capability of CLIP.\\
\textbf{Prompt.} Prompt learning is an emerging topic in computer vision and natural language processing. Previous studies commonly involve inserting learnable tokens into the model input. They add learnable embeddings to the text input\cite{coop,cocoop,textsimage,prefix,ptuning}, or incorporate them into the image input\cite{VPT,VP,register}. 
Other methods incorporate learnable tokens into both the text and image inputs\cite{apollo,multitask,maple,unified}. 
Most of these methods require retraining because they involve fine-tuning specific parameters of a pre-trained model to adapt to specific downstream task datasets.
Some works manually introduce prompts (box, circle, blur mask) within the images to guide the model towards the desired objects or regions\cite{CPT,reclip,redcircle,FGVP}. 
However, the majority of these methods are reliant on the pretraining data of CLIP, and they alter the original information of the images, making it difficult to generalize to certain downstream tasks. 
Our method can be regarded as a form of \textit{attention prompt}. What sets our method apart from these works is that our method does not need training, introduction of additional models, or altering the content of the original images. \\
\textbf{CLIP Region Awareness.} To enhance the region awareness of CLIP, several methods have been explored in the field of detection and segmentation. SAN\cite{SAN} trains a extra transformer network to assist CLIP in recognizing local features. ODISE\cite{ODISE} employ a trainable mask generator to guide CLIP's focus towards specific local regions of interest. RegionCLIP\cite{regionclip}, OvarNet\cite{ovarnet}, Alpha-CLIP\cite{alpha-clip} and UMG-CLIP\cite{UMG-CLIP} use region-level image-text pairs to fine-tune the model. MaskAdaptedCLIP\cite{maskadaptiveclip} generates mask-text pairs through a pseudo-labeling process to fine-tune CLIP. MaskCLIP\cite{maskclip} and MasQCLIP\cite{masqclip} introduce additional learnable tokens enhancing CLIP's ability to classify objects. Unlike our method, these methods have higher requirements for training data and often require additional training or fine-tuning processes.

\section{Method}
This section begins with a brief introduction to CLIP and visual prompts. We then proceed to discuss how to apply FALIP to zero-shot tasks such as referring expressions comprehension, image classification and 3D point cloud recognition. 

\begin{figure}[t]
\centering
\includegraphics[scale=0.7]{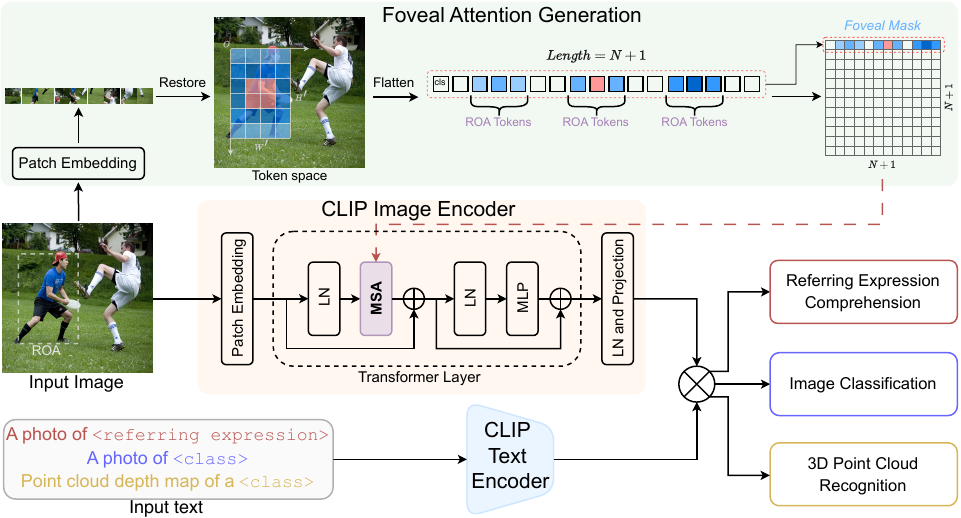}
\caption{FALIP Overview. We first input the image into the \textit{foveal attention generation} module to obtain a foveal attention mask. Then, we input original images to the CLIP image encoder, while also providing the foveal attention mask to the Multi-head Self-Attention (MSA) module. With different input images and text prompts, the model can accomplish tasks such as  referring expression comprehension, image classification and 3D point cloud recognition.
}
\label{fig:10}
\end{figure}

\subsection{Preliminary}

The core architecture of CLIP encompasses an image encoder $\mathbf{V}$ and a text encoder $\mathbf{T}$.
CLIP utilizes contrastive learning to distinguish between matching and mismatched image-text pairs. The text encoder is based on a Transformer\cite{transformer} architecture, while the image encoder can be either a ViT\cite{ViT} or a ResNet\cite{resnet}; our work utilizes the ViT model denoted as $\mathbf{V}$. 

When applying different visual prompts to images processed by CLIP's image encoder $\mathbf{V}$, we change the model's attention towards prompted regions, as indicated in \cref{fig:2} by the brightness of tokens in the multi-head self-attention. 
This observation suggests that visual prompts significantly influence the model's focus on specific image areas.
Based on our finding, we propose a hypothesis that the effectiveness of visual prompts can be fundamentally attributed to their ability to alter the model's attention. \cref{fig:10} presents the overall framework of our method.

\subsection{Foveal Attention}
\label{sec:foveal}
In a typical ViT network, an input image $\mathbf{x}\in \mathbb{R}^{C\times H\times W}$ will be processed as $N$ tokens $x_1,x_2,x_3,\cdot  \cdot  \cdot,x_n$. Denoting $x_{cls}$ as the [CLS] token, the input of the transformer layer can be represented as: $X = \{x_{cls},x_1,x_2,x_3,\cdot  \cdot \cdot,x_n\}\in \mathbb{R}^{(N+1)\times D}$. 
Our method takes both an image $\mathbf{x}\in \mathbb{R}^{C\times H\times W}$ and its corresponding attention mask $M\in\mathbb{R}^{(N+1)\times(N+1)}$ as inputs. Removing the [CLS] token and restoring the spatial position of $X$, 
we can identify the tokens that are originally located on the region of attention (ROA). We represent these tokens as $TOKEN_{roa}=\{x_n|x_n\ \text{located on the ROA}\}$.

To generate foveal attention mask for ROA, we first compute: 
\begin{equation}
R_{i,j} = e^{-\frac{[i-(H'-1)/2]^2+[j-(W'-1)/2]^2}{2\sigma^2}}
\end{equation}

\begin{equation}
R^{norm} = \alpha \times \frac{R-\text{Min}(R)+\epsilon}{\text{Max}(R)-\text{Min}(R)+\epsilon}
\end{equation}
where $\sigma$ and $\alpha$ are adjustable parameters, $\epsilon$ is a small constant, $H'\in[1,\sqrt{N}]$, $W'\in[1,\sqrt{N}]$ are height and width of ROA in token space. Flattening $R^{norm}$ and aligning its indices with $X$, the formula for $M$ is given as follows:

\begin{equation}
\ M_{i,j}=\left\{
\begin{array}{lcl}
R^{norm}_j        &        &\quad{x_j\in TOKEN_{roa}}\quad{and}\quad{i=0}\\
0        &        &\quad{x_j\notin TOKEN_{roa}}\quad{and}\quad{i=0}  \\
0        &        &\quad{i>0}
\end{array} \right.
\end{equation}
only the first row of $M$ is assigned non-zero value, we will discuss it in \cref{sec:ablation}. The formula for foveal attention is as follows:
\begin{equation}
\text{Foveal-Attention}(Q, K, V) = \text{Softmax} \left( \frac{QK^\mathsf{T}}{\sqrt{d}}+M \right) V
\end{equation}
The design of FALIP is inspired by the foveal characteristics in human visual attention. We introduce a gradual blending of the foveal mask to promote smooth transitions between focal regions and surrounding backgrounds. The mask assigns \textit{Gaussian-weighted} coefficients to the attentive tokens, mitigating the interference between background elements and focal regions. 


\subsection{Applications}
Now that we have introduced the main principle of FALIP, we proceed to deploy the augmented model on several zero-shot tasks and discuss detailed considerations specific to each task.

\subsubsection{Referring Expression Comprehension.}

Referring expression comprehension (REC) involves identifying an object in an image based on a textual description that explicitly refers to it. The entire process in a zero-shot manner can be represented as follows. The data for pre-processing includes an image $\mathbf{x}\in \mathbb{R}^{C\times H\times W}$, $B$ boxes 
and a text $t$. Based on the aforementioned conclusions,  $B$ boxes can be transformed into masks ${M^*}\in\mathbb{R}^{B\times{(N+1)^2}}=\{M_1,M_2,M_3,\cdot  \cdot \cdot,M_L\}$. The similarity between a text and a box region in the image and can be represented as follows:
$S_i = \mathbf{T}(t)\cdot\mathbf{V}^{\mathsf{T}}(\mathbf{x}, M_i)\quad i\in[1,B]$,
where `` $\cdot$ '' represents matrix multiplication.
Similar to previous work\cite{redcircle}, the ``subtract'' operation is utilized in the post-processing step to weigh down $S_i$. The best matching mask (box region) $M_k$ to $t$ is given by: $k = \mathop{\text{argmax}}\limits_{i} \left[S_i-\frac{1}{Q}\sum_{q=1}^{Q}\mathbf{T}(\hat{t}_q)\cdot\mathbf{V}^{\mathsf{T}}(\mathbf{x},M_i)\right]\quad \hat{t}\in\/\hat{T}$, where ${\hat{T}} = \{\hat{t}_1,\hat{t}_2,\hat{t}_3\,\cdot\cdot\cdot,\hat{t}_Q\}$. $\hat{T}$ is abtained by randomly sampling $Q$ negative captions related to no instances on the image from the whole dataset.

\subsubsection{Image Classification.}
In this application scenario, the inputs to FALIP include $\mathbf{x}\in \mathbb{R}^{C\times H\times W}$, texts $\widetilde{T}=\{\tilde{t_1},\tilde{t_2},\tilde{t_3},\cdot\cdot\cdot,\tilde{t_c}\}$, and boxes. Transforming boxes to $M$ the entire classification process formulation is as follows:
$Score_i = \mathbf{V}(\mathbf{x},M)\cdot\mathbf{T}^{\mathsf{T}}(\tilde{t_i})\quad i\in[1,c]$, $Pred = \mathop{\text{argmax}}\limits_i \left[\frac{e^{Score_i}}{\sum_{i=1}^{c}{e^{Score_i}}}\right]$, where $Pred$ is the index of the text corresponding to the image category in $\widetilde{T}$.
This task differs from REC, as there is only one mask as input in image classification task. 
%

\subsubsection{3D Point Cloud Recognition.} CLIP can be deployed for 3D point cloud recognition\cite{pointclip} by projecting a 3D point cloud into six views of 2D depth maps $\overline{\mathbf{x}}\in \mathbb{R}^{6\times C\times H\times W}$. We locate the foreground positions in the depth maps and convert them into ${M^*}\in\mathbb{R}^{6\times{(N+1)^2}}=\{M_1,M_2,\cdot  \cdot \cdot,M_6\}$. The texts of the category is $\overline{T}=\{\overline{t}_1,\overline{t}_2,\overline{t}_3,\cdot\cdot\cdot,\overline{t}_c\}$.
The recognition process with FALIP is as follows: $Score_i = \sum_{j=1}^{6}\beta_j\mathbf{V}(\overline{\mathbf{x}}_j,M_j)\cdot\mathbf{T}^{\mathsf{T}}(\overline{t}_i)\quad i\in[1,c]$, $Pred = \mathop{\text{argmax}}\limits_i \left[\frac{e^{Score_i}}{\sum_{i=1}^{c}{e^{Score_i}}}\right]$, where $\beta$ is used to control the weights of views. $Pred$ is the index of the text corresponding to the image category in $\overline{T}$.

\section{Experiments}

In this section, we begin by comparing our method with existing visual prompt methods on the referring expression comprehension task. Lastly, we extend our method to other tasks like image classification and 3D point cloud recognition, showcasing its superiority. Finally, we present our observations on the visual prompts and introduce the ablation experiments of FALIP. Unless otherwise specified, our experiments are conducted using the OpenAI version of the ViT/B-16 CLIP model. All experiments are performed on two RTX 3090 GPUs. For more experimental details, please refer to the Appendix.


\subsection{Referring Expression Comprehension}
We conduct the REC task on the RefCOCO\cite{refcoco}, RefCOCO+\cite{refcoco}, and RefCOCOg\cite{refcocog} datasets. RefCOCO+ focuses on appearance-based expressions, while RefCOCO and RefCOCOg include relation-based expressions. The test sets are divided into two subsets: TestA (expressions referring to people) and TestB (expressions referring to non-people objects).

\begin{figure}[t]
\centering
\includegraphics[scale=0.57]{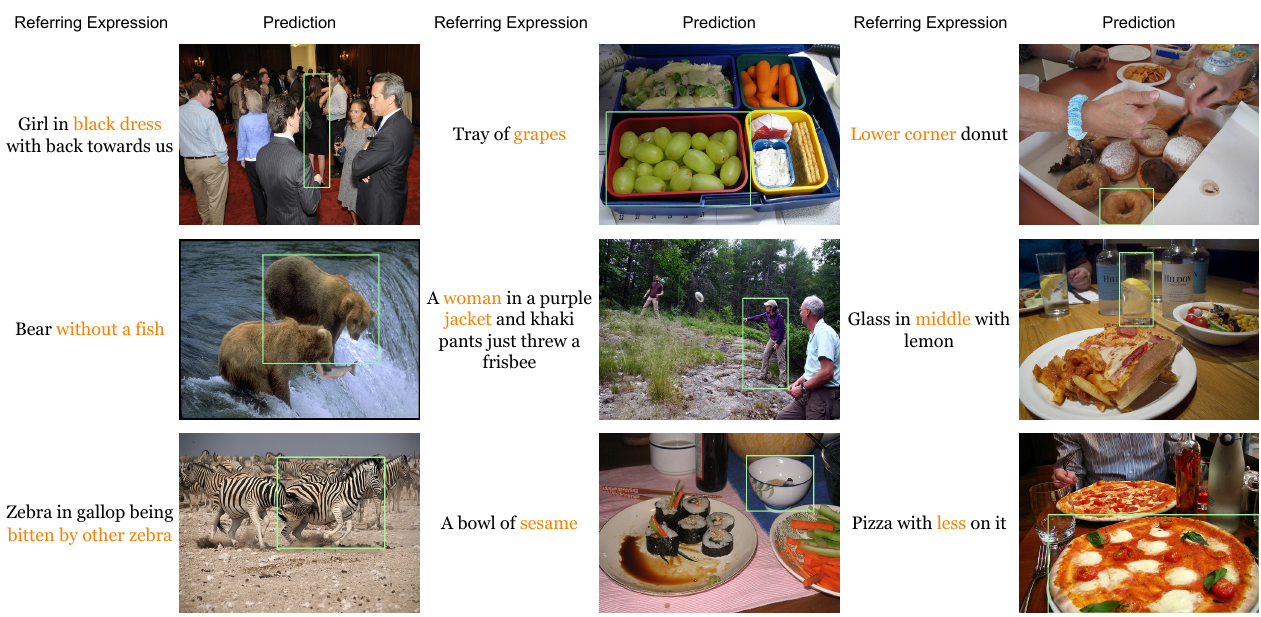}
\caption{Visualization of referring expression comprehension. The model predicts the corresponding object in the image based on the given referring expression. The key words in referring expression is colored \textcolor{orange}{orange}.
}
\label{fig:9}
\end{figure}

In previous works\cite{mattnet,uniter}, some methods first extract proposals from the image using object detectors or instance detectors\cite{fasterrcnn,maskrcnn}, and then score the matching degree between these bounding boxes and the given text. To simplify this process and alleviate the need for explicit proposals, methods such as ViLT\cite{vilt} and other methods\cite{mdetr,coarsetofine,referring} adopt an end-to-end training approach to predict a bounding box corresponding to the referring expression. We compare our method FALIP with previous zero-shot methods\cite{CPT,redcircle,reclip} in \cref{table:rec}. Except for CPT\cite{CPT} using VinVL\cite{vinvl} model, all others use the ViT-B model. We conducte experiments using two bounding box settings ``prop'' and ``gold'', which respectively represent the proposals generated by the MAttNet\cite{mattnet} and the annotations from the datasets. The results under the ``gold'' setting are generally better than those under the ``prop'' setting, indicating that using a more powerful detector or applying  filtering to the proposals can lead to improved zero-shot accuracy. Our method outperforms existing methods in setting ``Without E and P''. Additionally, it can be combined with existing methods to further enhance their performance. When ``subtract'' post-progressing is utilized, FALIP shows a significant improvement in accuracy, but it has minimal impact on the performance for the RedCircle.  FALIP maintains competitive performance when used in conjunction with ensemble and post-processing, surpassing existing methods by approximately 3\%. The results demonstrate that our method is effective in various settings. Visualization results can be seen in \cref{fig:9}. \\

\renewcommand{\arraystretch}{0.82}

\setlength{\tabcolsep}{5.5pt}
\begin{table}[t]
\begin{center}
\caption{Results on Referring Expressions Comprehension. ``FA'': Foveal-attention. ``P'': Post-progressing. ``E'': Ensemble-prompt. ``$\ddagger$'': Results from the original paper. Our method is effective in this task. The best results are in \textbf{bold}.}
\label{table:rec}

\begin{tabular}{lcccccccc}
\toprule
\multirow{2}{*}{Method} & \multicolumn{3}{c}{RefCOCO} & \multicolumn{3}{c}{RefCOCO+} & \multicolumn{2}{c}{RefCOCOg}\\ 
& TestA & TestB & Val & TestA & TestB & Val & Test & Val \\
\midrule
Without E and P \\
\midrule
CPT\cite{CPT}$\ddagger$$_{prop}$ & 36.1 & 30.3 & 32.2 & 35.2 & 28.8 & 31.9 & 36.5 & 36.7 \\
RedCircle\cite{redcircle}$_{prop}$ & 38.8 & 30.5 & 34.9 & 41.7 & 31.9 & 37.7 & 39.7 & 39.7 \\
PASTA\cite{PASTA}$_{prop}$ & 39.3 & 32.7 & 36.3 & 40.4 & 36.2 & 38.4 & 43.8 & 43.8 \\
RedCircle+FA$_{prop}$ & 40.7 & 31.7 & 35.9 & 43.6 & 34.0 & 39.2 & 41.0 & 41.6 \\
\rowcolor{violet!10}
FALIP(Ours)$_{prop}$ & 41.4 & 33.2 & 37.5 & 44.4 & 37.6 & 40.3 & 45.4 & 45.6 \\
RedCircle\cite{redcircle}$_{gold}$ & 41.6 & 36.2 & 38.2 & 44.7 & 37.7 & 41.1 & 45.4 & 45.7 \\
PASTA\cite{PASTA}$_{gold}$ & 41.7 & 37.6 & 39.5 & 43.2 & 40.5 & 42.4 & 49.2 & 49.9 \\
RedCircle+FA$_{gold}$ & 41.8 & 36.9 & 38.2 & 45.1 & 38.4 & 41.6 & 46.1 & 46.1 \\
\rowcolor{violet!10}
FALIP(Ours)$_{gold}$ & \textbf{44.2} & \textbf{39.4} & \textbf{40.8} & \textbf{46.8} & \textbf{43.1} & \textbf{44.5} & \textbf{51.5} & \textbf{51.3} \\
\midrule
With P \\
\midrule
RedCircle$_{prop}$ & 34.3 & 30.3 & 33.8 & 36.8 & 31.0 & 36.3 & 39.1 & 39.2 \\
\rowcolor{violet!10}
FALIP(Ours)$_{prop}$ & 49.0 & 39.1 & 44.7 & 52.5 & 43.0 & 48.2 & 51.3 & 51.6 \\
RedCircle$_{gold}$ & 39.2 & 37.7 & 39.5 & 42.8 & 39.7 & 42.2 & 44.9 & 45.3 \\
\rowcolor{violet!10}
FALIP(Ours)$_{gold}$ & \textbf{50.6} & \textbf{44.8} & \textbf{48.3} & \textbf{54.5} & \textbf{48.6} & \textbf{52.0} & \textbf{57.1} & \textbf{56.9} \\
\midrule
With E and P \\
\midrule
ReCLIP\cite{reclip}$_{prop}$ & 42.4 & 44.4 & 42.1 & 42.8 & 42.3 & 41.9 & 56.9 & 57.1 \\
RedCircle$\ddagger$$_{prop}$ & 52.7 & 36.5 & 45.3 & 57.7 & 40.6 & 49.4 & 53.3 & 53.7 \\
\rowcolor{violet!10}
FALIP(Ours)$_{prop}$ & 51.7 & 38.3 & 46.7 & 57.1 & 43.0 & 51.9 & 54.9 & 54.2 \\
RedCircle$_{gold}$ & \textbf{53.5} & 41.7 & 46.9 & 58.5 & 45.5 & 52.3 & 57.3 & 56.9 \\
\rowcolor{violet!10}
FALIP(Ours)$_{gold}$ & 53.4 & \textbf{44.6} & \textbf{49.8} & \textbf{59.4} & \textbf{49.5} & \textbf{55.1} & \textbf{60.7} & \textbf{59.3} \\
\bottomrule
\end{tabular}
\end{center}
\end{table}

\begin{figure}[t]
\centering
\includegraphics[scale=0.55]{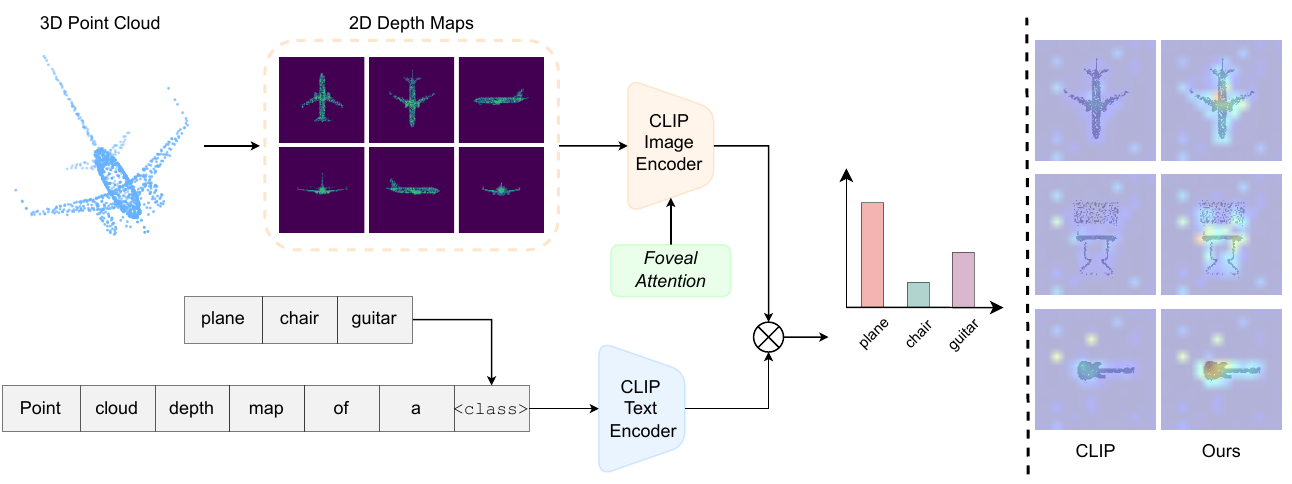}
\caption{Pipeline of 3D point cloud recognition. \textit{\textbf{Left}}: The overall framework remains consistent with PointCLIP\cite{pointclip}, with the difference being the insertion of foveal attention in the CLIP image encoder. \textit{\textbf{Right}}: Attention on the 2D depth maps of original CLIP and our method. It can be observed that our method shows a stronger attention towards the foreground.
}
\label{fig:6}
\end{figure}

\setlength{\tabcolsep}{7pt}
\begin{table}[t]
\begin{center}
\caption{Results on Image Classification. ``Blur'' refers to applying blur operation to the areas outside the circle. The superiority of our method lies in preserving the original fine-grained features of the image. The best results are in \textbf{bold}, and sub-optimal results are \underline{underlined}.}
\label{table:classification}
\begin{tabular}{lcccccccc}
\toprule
\multirow{2}{*}{Method} & \multicolumn{2}{c}{StanfordDogs} & \multicolumn{2}{c}{CUB-200-2011} & \multicolumn{2}{c}{ImageNet-S} & \ {Waterbirds} \\
& Top1 & Top5 & Top1 & Top5 & Top1 & Top5 & Top1 \\
\midrule
Original CLIP & \underline{56.5}& \underline{85.2} & \underline{54.2} & \textbf{83.7} & \underline{64.9} & \underline{88.4} & \underline{78.2} \\
RedCircle & 52.4 & 82.8 & 44.2 & 77.0 & 62.8 & 86.5 & 77.5\\
Blur & 51.9 & 81.9 & 39.1 & 71.0 &53.8 &77.6 & 78.1\\
\rowcolor{violet!10}
FALIP(Ours) & \textbf{58.3}& \textbf{86.0} & \textbf{54.3} & \underline{83.6}  & \textbf{67.3} &\textbf{89.9} & \textbf{79.7}\\

\bottomrule
\end{tabular}
\end{center}
\end{table}

\subsection{Image Classification}
For classification, we use the StanfordDogs\cite{stanforddogs}, CUB-200-2011\cite{cub200}, Waterbirds\cite{waterbirds}, and ImageNet-S\cite{imagenet-s} datasets. StanfordDogs and CUB-200-2011 consist of images of 120 different dog breeds and 200 bird species, respectively. Waterbirds contains photographs of waterbirds and landbirds, with bird images from the CUB dataset and backgrounds from the Places dataset\cite{places}. Notably, Waterbirds is a binary classification dataset. ImageNet-S includes 919 classes with semantic segmentation annotations, selected from ImageNet-1k\cite{imagenet}.

We compare our method FALIP with previous visual prompt and original CLIP in classification task in \cref{table:classification}. On the four classification datasets, FALIP improves classification accuracy, while the visual prompts RedCircle and Blur respectively leads to slight and significant decreases in accuracy. RedCircle can be seen as contamination to the original fine-grained features. Blur blurs out a significant portion of the background and some foreground subject features. This results in a significant accuracy decrease across the first three datasets. However, in the Waterbirds dataset, where the classification decision relies on the background, Blur eliminates the interfering factor, resulting in only a slight accuracy decrease. We believe that the failure of the visual prompt is due to the contamination introduced by altering the image content, which impacts the fine-grained image classification performance negatively.

\subsection{3D Point Cloud Recognition}
For 3D point cloud recognition, we use the ModelNet40\cite{modelnet40} and ScanObjectNN\cite{scanobjectnn} datasets. ModelNet40 has 40 object categories, encompassing common objects like chairs and airplanes. ScanObjectNN comprises real-world 3D scan data with 15 categories of household objects such as tables and lamps.

We extend the proposed FALIP to 3D point cloud recognition. In the experiments, we use PointCLIP\cite{pointclip} as the baseline, which employs the CLIP as the image encoder. The details are shown in \cref{fig:6}. The experimental results are presented in \cref{table:3d}. FALIP outperforms the original CLIP on both datasets, achieving an average accuracy improvement of 1.4\%. This encouraging result demonstrates FALIP's potential to extend to 3D data domains.

\setlength{\tabcolsep}{20pt}
\begin{table}[t]
\begin{center}
\caption{Results on 3D point cloud recognition. Our method improves the recognition capability of CLIP. The best results are in \textbf{bold}.}
\label{table:3d}
\begin{tabular}{lcccccccccc}
\toprule
{Method} & {ModelNet40} & {ScanObjectNN} & {Avg}\\ 
\midrule
Original CLIP & 16.5 & 14.6 & 15.6 \\
\rowcolor{violet!10}FALIP(Ours) & \textbf{18.6} & \textbf{15.3} & \textbf{17.0} \\

\bottomrule
\end{tabular}
\end{center}
\end{table}

\setlength{\tabcolsep}{5.5pt}
\begin{table}[t]
\begin{center}
\caption{Ablation on unleashing of viusal prompt on REC task. ``R'': RedCircle. ``B'': Blur. ``U'': Unleash. Adjusting salient attention heads increases the concentration of the model's attention, since unleashed visual prompt outperforms original visual prompt on all metrics. Details are in \cref{sec:unleash}. The best results are in \textbf{bold}. }
\label{table:unleash}
\begin{tabular}{ccccccccccccc}
\toprule
\multirow{2}{*}{R} & \multirow{2}{*}{B} & \multirow{2}{*}{U} & \multicolumn{3}{c}{RefCOCO} & \multicolumn{3}{c}{RefCOCO+} & \multicolumn{2}{c}{RefCOCOg} & \multirow{2}{*}{Avg}\\ 
& & & TestA & TestB & Val & TestA & TestB & Val & Test & Val \\
\midrule
\checkmark & & & 41.6 & 36.2 & 38.2 & 44.7 & 37.7 & 41.1 & 45.4 & 45.7 & 41.3 \\
\checkmark & & \checkmark & \textbf{46.1} & \textbf{39.4} & \textbf{42.0} & \textbf{50.1} & \textbf{40.3} & \textbf{44.8} & \textbf{49.9} & \textbf{49.4} & \textbf{45.2} \\
\midrule
\checkmark & \checkmark & & 45.4 & 41.5 & 43.8 & 49.3 & 44.9 & 46.5 & 56.6 & 56.9 & 48.3 \\
\checkmark & \checkmark & \checkmark & \textbf{49.0} & \textbf{41.6} & \textbf{45.7} & \textbf{54.6} & \textbf{45.1} & \textbf{49.7} & \textbf{56.7} & \textbf{57.0} & \textbf{49.9} \\

\bottomrule
\end{tabular}
\end{center}
\end{table}

\begin{figure}[t]
\centering
\includegraphics[scale=0.525]{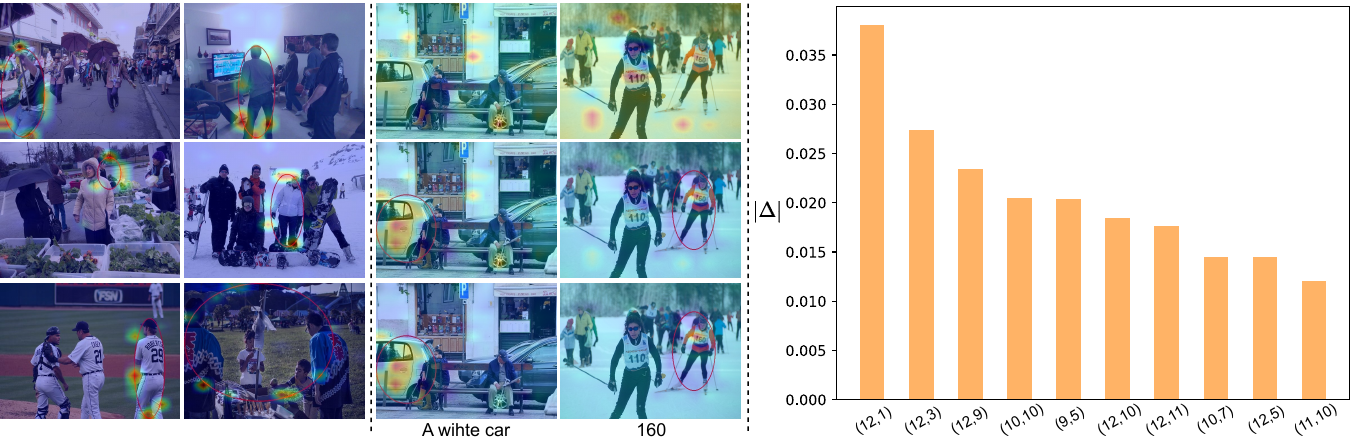}
\caption{Visualization of attention. The attention maps are generated by\cite{interpreting}. {\it \textbf{Left}}: Attention maps of Layer9-Head5 in CLIP image encoder for various images. This attention head shows sensitivity towards RedCircles of varying positions and sizes. {\it \textbf{Middle}}: From top to bottom, are the original image, the image with RedCircle, and the image with unleashed RedCircle. The attention maps for the text input show a gradual decrease in attention on irrelevant backgrounds. {\it \textbf{Right}}: Ranking effect on attention head labeled by
(Layer, Head).
}
\label{fig:5}
\end{figure}

\subsection{Unleash Visual Prompts}
\label{sec:unleash}
To investigate the impact of visual prompts on each attention head, we decouple the output of the image encoder $\mathbf{V}$ into the sum of the individual attention heads. The specific steps are as follows: $X_l' = \text{MSA}[\text{LN}(X_{l-1})]+X_{l-1}$, $X_l = \text{MLP}[\text{LN}(X_l')]+X_l'$, where $l\in[1,L]$, $L$ represents the total layers of model. $\text{MSA}$ is multi-head self-attention. $\text{LN}$ is LayerNorm operation, $X_l$ is the output of the $l$-th layer. Denoting $[X_l]_{cls}$ as [CLS] token in $X_l$, the output of $\mathbf{V}$ can be expressed as:
\begin{equation}
[X_L]_{cls} = [X_0]_{cls}+\sum_{l=1}^{L}\Bigl[\text{MSA}[\text{LN}(X_{l-1})]\Bigr]_{cls}+\sum_{l=1}^{L}\Bigl[\text{MLP}[\text{LN}(X_l')]\Bigr]_{cls} \label{eq:11}
\end{equation}
Here we omit the final project layer, and decouple the final output into the sum of $\text{MSA}$ and $\text{MLP}$. Further decomposition of the second term in \cref{eq:11}: $\Bigl[\text{MSA}[\text{LN}(X_{l-1})]\Bigr]_{cls} = \sum_{h=1}^{H}\sum_{i=1}^{N+1}p_{i,h}\quad p_{i,h}= \gamma_i^h\text{LN}(x_i^{l-1})W_V^h\label{eq:12}$, where $H$ is number of attention heads, $N+1$ is the number of input tokens. $\gamma_i^h$ is the attention weights between [CLS] token and $i$-th token. $x_i^{l-1}$ is $i$-th token in $X^{l-1}$. $W_V^h$ is a mapping matrix of $V$. Let $G_h=\sum_{i=1}^{N+1}p_{i,h}$, the change of MSA [CLS] token after using visual prompt is: $\Delta=\Bigl[\text{MSA}[\text{LN}(X_{l-1})]\Bigr]_{cls}'-\Bigl[\text{MSA}[\text{LN}(X_{l-1})]\Bigr]_{cls}= \sum_{h=1}^{H}(G'_h-G_h)\label{eq:13}$. Thus, we can easily observe the changes in individual attention heads across different layers before and after using the visual prompt. 

We compute the changes in individual attention heads before and after using RedCircle on a large number of images and find that the attention heads in the last 4 layers of the model exhibited significant variations (shown in \cref{fig:5}). Therefore, we propose generating a new [CLS] token by editing these self-attention heads. The formula for the new $\text{MSA}$ [CLS] token is as follows: 
\begin{equation}
\Bigl[\text{MSA}[\text{LN}(X_{l-1})]\Bigr]_{cls}= \sum_{h=1}^{H}[G'_h+(G'_h-G_h)] \quad\quad l\in[L-3,L]\label{eq:14}
\end{equation}

By substituting \cref{eq:14} into \cref{eq:11}, we can obtain the new output $[X_L]_{cls}'$ of the model. We test the generated new output on the REC task in \cref{table:unleash}. After unleashing the potential of RedCircle, the average accuracy has increased by more than 4\%. By applying the same method to RedCircle+Blur, an improvement in performance can be seen as well. This phenomenon suggests that the potential of the current visual prompt has not been fully explored. Furthermore, we discover that some attention heads in CLIP are particularly sensitive to RedCircle. \cref{fig:5} demonstrates some visualizations to explain these.

\setlength{\tabcolsep}{5pt}
\begin{table}[t]
\begin{center}
\caption{Ablation on different $q$, $k$, $v$ in self-attention. The results in the second row and the third row correspond to the methods illustrated in \cref{fig:7}b and \cref{fig:7}c, respectively. The best results are in \textbf{bold}.}
\label{table:qkv}
\begin{tabular}{lcccccccccc}
\toprule
\multirow{2}{*}{Method} & \multicolumn{3}{c}{RefCOCO} & \multicolumn{3}{c}{RefCOCO+} & \multicolumn{2}{c}{RefCOCOg} & \multirow{2}{*}{Avg}\\ 
& TestA & TestB & Val & TestA & TestB & Val & Test & Val \\
\midrule
RedCircle & 41.6 & 36.2 & 38.2 & 44.7 & 37.7 & 41.1 & 45.4 & 45.7 & 41.3 \\
Replace $v$ & 38.5 & 35.1 & 36.2 & 40.6 & 37.1 & 38.3 & 43.4 & 42.6 & 39.0 \\
Replace $q,k$ & 39.9 & 32.9 & 35.9 & 42.1 & 35.3 & 38.3 & 41.7 & 42.4 & 38.6 \\
\midrule
Feature mask & 25.4 & 26.4 & 25.6 & 24.9 & 27.7 & 26.2 & 29.8 & 30.0 & 27.0 \\
\rowcolor{violet!10}
FALIP(Ours) & \textbf{44.2} & \textbf{39.4} & \textbf{40.8} & \textbf{46.8} & \textbf{43.1} & \textbf{44.5} & \textbf{51.5} & \textbf{51.3} & \textbf{45.2} \\
\bottomrule
\end{tabular}
\end{center}
\end{table}

\begin{figure}[t]
\centering
\includegraphics[scale=0.75]{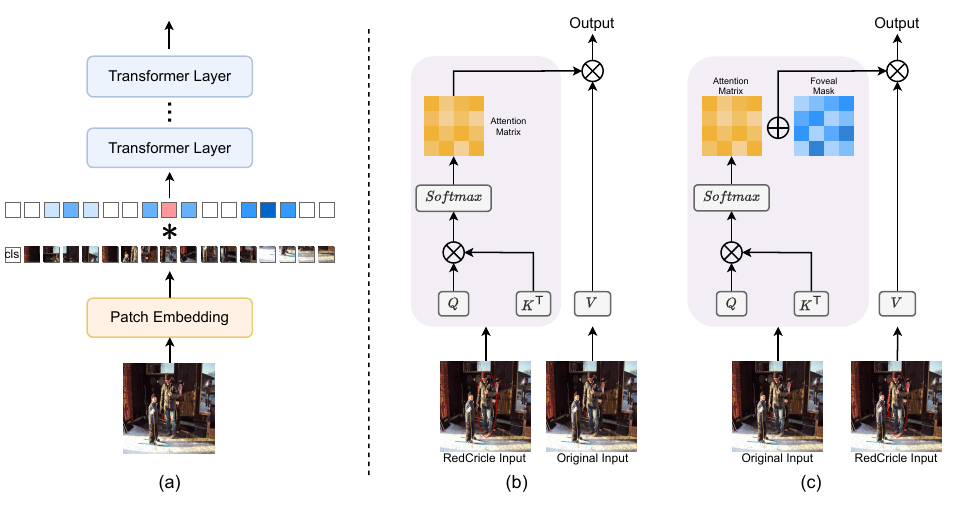}
\caption{Illustration of baseline methods in Table~\ref{table:qkv}. (a) Feature mask. ``$*$'' means element-wise multiplication. (b) Self-attention using the $q$, $k$ generated from the RedCircle image and the $v$ from the original image. (c) Self-attention using $q$, $k$ generated from the original image and $v$ from the RedCircle image with foveal attention mask.
}
\label{fig:7}
\end{figure}

\begin{figure}[t]
\centering
\includegraphics[scale=0.6]{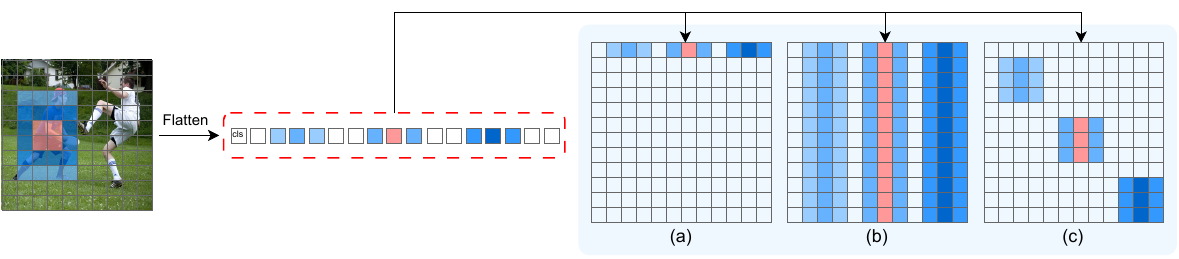}
\caption{Several ways to generate foveal attention masks. (a) Only assigning values at the position corresponding to the ROA token in the first row. (b) Assigning values of ROA token position in all rows. (c) Assigning values at the position corresponding to the ROA token on the diagonal line.
}
\label{fig:3}
\end{figure}

\subsection{Ablation on Foveal Attention}
\label{sec:ablation}
\subsubsection{Other implementations.} \cref{fig:7}a shows feature mask method, the results can be seem in \cref{table:qkv}. The accuracy of feature mask is significantly lower compared to our method. \cref{fig:7}b and \cref{fig:7}c illustrate two methods for performing self-attention by replacing $q$, $k$, and $v$. We observe that regardless of whether $q$, $k$, or $v$ is replaced, the accuracy of the RedCircle decreases. This suggests that the $q$, $k$, and $v$ generated from the images with trained visual prompts have a strong correlation. The details is shown in \cref{table:qkv}.
\subsubsection{Forms of masks.} As in \cref{fig:3}, we use three different forms of masks. The results in \cref{table:mask} indicate that \textit{Method a} achieve the highest accuracy. \textit{Method b} causes all other tokens to pay excessive attention to the specified region, disrupting the original information carried by these tokens. On the other hand, \textit{Method c} causes an excessive focus on the tokens within the specified region, neglecting the contextual information. This suggests the row corresponding to [CLS] token plays a crucial role in the model's prediction outcomes.

\setlength{\tabcolsep}{5.7pt}
\begin{table}[t]
\begin{center}
\caption{Ablation on the forms of masks. ``No mask'' means original CLIP. ``Method a'' is the optimal form of the mask, which preserve the original information carried by tokens, and reassign the weights of all tokens with respect to the [CLS] token. The best results are in \textbf{bold}.}
\label{table:mask}
\begin{tabular}{lcccccccccc}
\toprule
\multirow{2}{*}{Method} & \multicolumn{3}{c}{RefCOCO} & \multicolumn{3}{c}{RefCOCO+} & \multicolumn{2}{c}{RefCOCOg} & \multirow{2}{*}{Avg}\\ 
& TestA & TestB & Val & TestA & TestB & Val & Test & Val \\
\midrule
No mask& 14.8 & 25.5 & 19.5 & 14.7 & 26.1 & 19.8 & 25.5 & 26.3 & 21.5 \\
Method a & \textbf{44.2} & \textbf{39.4} & \textbf{40.8} & \textbf{46.8} & \textbf{43.1} & \textbf{44.5} & \textbf{51.5} & \textbf{51.3} & \textbf{45.2} \\
Method b & 36.6 & 37.2 & 35.1 & 39.1 & 40.3 & 37.8 & 43.7 & 44.2 & 39.3 \\
Method c & 13.3 & 19.0 & 15.9 & 12.5 & 19.0 & 15.1 & 16.1 & 15.6 & 15.8 \\

\bottomrule
\end{tabular}
\end{center}
\end{table}

\begin{figure}[t]
\centering
\includegraphics[scale=0.58]{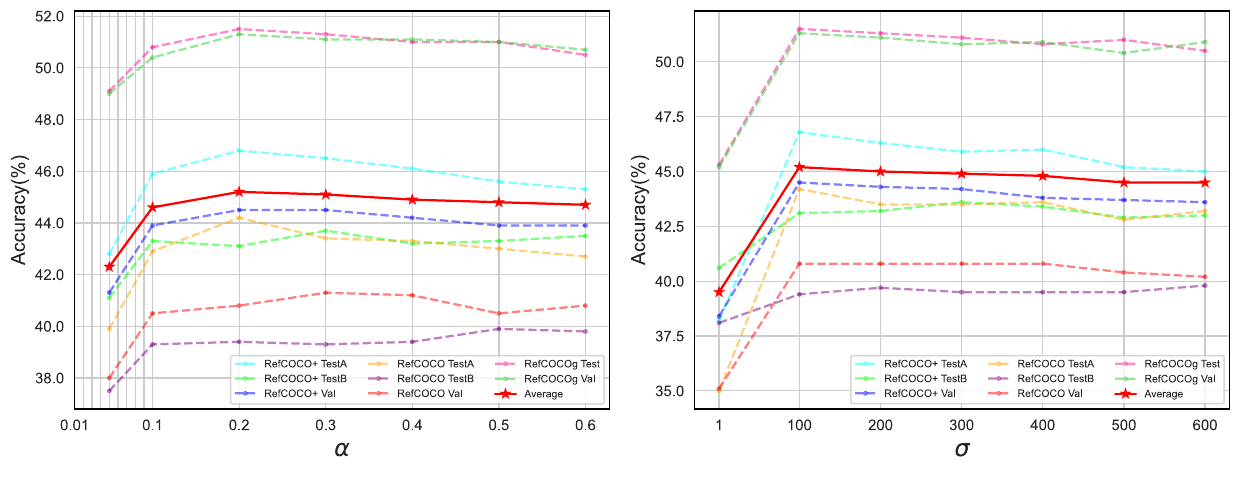}
\caption{Effect of $\alpha$ and $\sigma$ in masks. 
The accuracy peaks when $\alpha=0.2$, as it highlights the features of the specific object while preserving the contextual information. When $\sigma=1$, the concentration of values within a mask can lead to a lack of rich features in specific regions. We consider $\alpha=0.2$ and $\sigma=100$ as the optimal value.
}
\label{fig:4}
\end{figure}



\subsubsection{Value of $\mathbf{\alpha}$ and $\mathbf{\sigma}$.} We conduct ablation experiments on the value of $\alpha$, $\sigma$ defined in \cref{sec:foveal} and present results in \cref{fig:4}. 


\section{Conclusion}
This paper presents an exploration into the surprising effectiveness of visual prompts for CLIP. We discover a close relationship between visual prompts and the attention mechanism within CLIP. Motivated by this finding, we propose FALIP, which enhances the region-awareness capability of CLIP by mimicking human attention characteristics, without incurring additional model fine-tuning or sacrificing its pre-trained knowledge. Our method achieves state-of-the-art performance on the referring expression comprehension task and demonstrates significant improvements on tasks like image classification and 3D point cloud recognition. Furthermore, we discover that the full potential of visual prompts can be further unleashed by adjusting the salient attention heads. We hope this work can provide inspiration for the understanding or design of visual prompts and attention mechanisms, sparking greater efforts from the research community dedicated to enhancing our understanding of intriguing phenomena exhibited by AI models.

\section*{Acknowledgments}
This work is supported by the National Natural Science Foundation of China (Grant No. U21B2004) and the Zhejiang Provincial Key RD Program of China (Grant No. 2021C01119).

\clearpage

%
%

\clearpage

\section*{Appendix}
\subsubsection{Datasets.} \cref{table:rec_data}, \cref{table:data_class} and \cref{table:data_3D} provide a brief introduction to the datasets used for tasks referring expression comprehension, image classification and 3D cloud recognition, respectively.

\setlength{\tabcolsep}{7pt}
\begin{table}
\begin{center}
\caption{Referring expression comprehension datasets. ``Refs'' means the number of referring expressions.}
\label{table:rec_data}
\begin{tabular}{lcccccccccc}
\toprule
&\multicolumn{3}{c}{RefCOCO} & \multicolumn{3}{c}{RefCOCO+} & \multicolumn{2}{c}{RefCOCOg} \\ 
& TestA & TestB & Val & TestA & TestB & Val & Test & Val \\
\midrule
Images & 750 & 750 & 1,500 & 750 & 750 & 1,500 & 2,600 & 1,300  \\
Refs & 1,975 & 1,810 & 3,811 & 1,975 & 1,798 & 3,805 & 5,023 & 2,573 \\

\bottomrule
\end{tabular}
\end{center}
\end{table}

\setlength{\tabcolsep}{7pt}
\begin{table}
\begin{center}
\caption{Image Classification datasets. ``Images used'' means the number of images used in our experiments.}
\label{table:data_class}
\begin{tabular}{lcccccccccc}

\toprule
&{StanfordDogs} & {CUB-200-2011} & {ImageNet-S} & \ {Waterbirds} \\
\midrule
Categories & 120 & 200 & 919 & 2\\ 
Total Images & 20,580 & 11,788  & 1,223,164 & 20,580\\
Images used & 20,580 & 11,788 & 12,419 & 5,794 & \\ 

\bottomrule
\end{tabular}
\end{center}
\end{table}

\setlength{\tabcolsep}{22pt}
\begin{table}
\begin{center}
\caption{3D cloud recognition datasets. ``Clouds used'' means the number of clouds used in our experiments.}
\label{table:data_3D}
\begin{tabular}{lcccccccccc}

\toprule
& ModelNet40 & ScanObjectNN & \\
\midrule
Categories & 40 & 15 \\ 
Total Clouds & 12,311 & 2,880 \\
Clouds used & 2,468 & 576 \\ 

\bottomrule
\end{tabular}
\end{center}
\end{table}

\subsubsection{Referring Expression Comprehension.} 
\begin{figure}[t]
\centering
\includegraphics[scale=0.75]{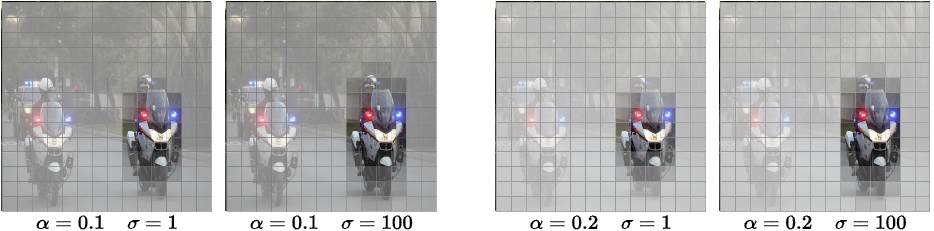}
\caption{Visualizing different values of $\alpha$ and $\sigma$ on the original image. A large $\alpha$ enhance prominence of the specific region and a large $\sigma$ preserve more content within the region.
}
\label{fig:8}
\end{figure}

\begin{figure}[t]
\centering
\includegraphics[scale=0.55]{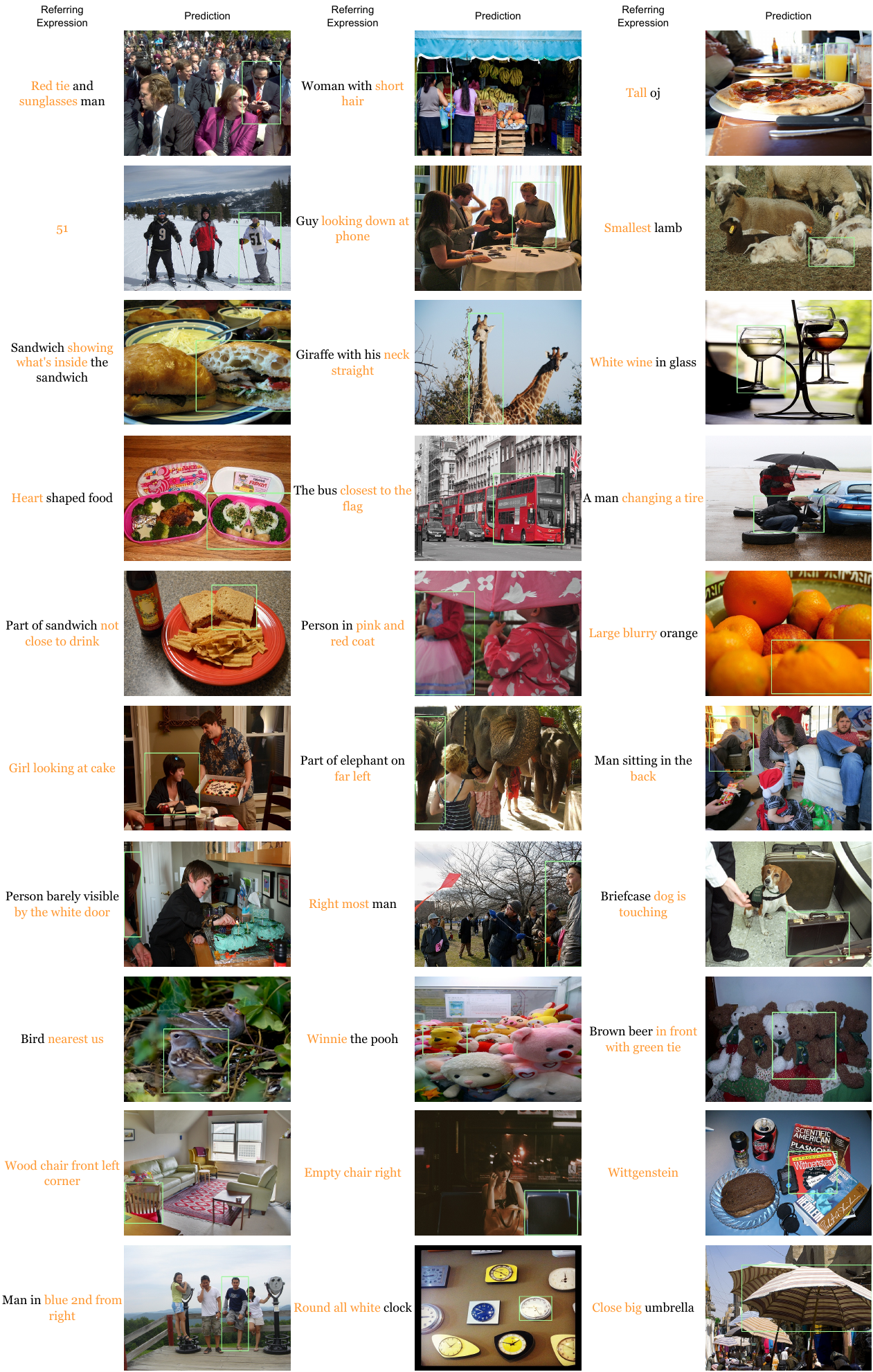}
\caption{The visualization results of REC. The keywords are highlighted in \textcolor{orange}{orange}.
}
\label{fig:a0}
\end{figure}

\cref{table:alpha} and \cref{table:sigma} present detailed experimental results about $\alpha$ and $\sigma$, respectively. We take $\alpha=0.2$ and $\sigma=100$ in final result. \cref{fig:8} illustrates the visual impact of different $\alpha$ and $\sigma$ on the original image. To investigate the sensitivity of different layers in CLIP to masks, we insert masks at various layers and present results in \cref{table:layer}. We find that inserting masks only in the last 4 layers results in the highest model accuracy, which suggests that the attention computations in the later layers play a decisive role in shaping the representation of the model's output, while the initial layers seem to have a minor impact on the results. \cref{fig:a2} depicts the details of the ensemble and \cref{fig:a0} shows the extensive results of referring expression comprehension. \\
\vspace{-0.3cm}

\setlength{\tabcolsep}{6.5pt}
\begin{table}
\begin{center}
\caption{Ablation on $\alpha$. The best results are in \textbf{bold}.}
\label{table:alpha}
\begin{tabular}{ccccccccccc}
\toprule
\multirow{2}{*}{$\alpha$} & \multicolumn{3}{c}{RefCOCO} & \multicolumn{3}{c}{RefCOCO+} & \multicolumn{2}{c}{RefCOCOg} & \multirow{2}{*}{Avg}\\ 
& TestA & TestB & Val & TestA & TestB & Val & Test & Val \\
\midrule
0.05 & 39.9 & 37.5 & 38.0 & 42.8 & 41.2 & 41.3 & 49.1 & 48.8 & 42.3 \\
0.1 & 42.9 & 39.3 & 40.5 & 45.9 & 43.3 & 43.9 & 50.8 & 50.4 & 44.6\\
\rowcolor{gray!25}
0.2 & \textbf{44.2} & 39.4 & 40.8 & \textbf{46.8} & 43.1 & \textbf{44.5} & \textbf{51.5} & \textbf{51.3} & \textbf{45.2} \\
0.3 & 43.4 & 39.3 & \textbf{41.3} & 46.5 & \textbf{43.7} & 44.5 & 51.3 & 51.1 & 45.1 \\
0.4 & 43.3 & 39.4 & 41.2 & 46.1 & 43.2 & 44.2 & 51.0 & 51.1 & 44.9 \\
0.5 & 43.0 & \textbf{39.9} & 40.5 & 45.6 & 43.3 & 44.0 & 51.0 & 51.0 & 44.8 \\
0.6 & 42.7 & 39.8 & 40.8 & 45.3 & 43.5 & 44.0 & 50.5 & 50.7 & 44.7 \\

\bottomrule
\end{tabular}
\vspace{-1.0cm}
\end{center}
\end{table}

\setlength{\tabcolsep}{6.5pt}
\begin{table}
\begin{center}
\caption{Ablation on $\sigma$. The best results are in \textbf{bold}.}
\label{table:sigma}
\begin{tabular}{ccccccccccc}
\toprule
\multirow{2}{*}{$\sigma$} & \multicolumn{3}{c}{RefCOCO} & \multicolumn{3}{c}{RefCOCO+} & \multicolumn{2}{c}{RefCOCOg} & \multirow{2}{*}{Avg}\\ 
& TestA & TestB & Val & TestA & TestB & Val & Test & Val \\
\midrule
1 & 35.0 & 38.1 & 35.1 & 38.2 & 40.6 & 38.4 & 45.3 & 45.2 & 39.5 \\
\rowcolor{gray!25}
100 & \textbf{44.2} & 39.4 & \textbf{40.8} & \textbf{46.8} & 43.1 & \textbf{44.5} & \textbf{51.5} & \textbf{51.3} & \textbf{45.2} \\
200 & 43.5 & 39.7 & 40.8 & 46.3 & 43.2 & 44.3 & 51.3 & 51.1 & 45.0 \\
300 & 43.5 & 39.5 & 40.8 & 45.9 & \textbf{43.6} & 44.2 & 51.0 & 50.8 & 44.9 \\
400 & 43.6 & 39.5 & 40.8 & 46.0 & 43.4 & 43.8 & 50.8 & 50.9 & 44.9 \\
500 & 42.8 & 39.5 & 40.4 & 45.2 & 42.9 & 43.7 & 51.0 & 50.4 & 44.5 \\
600 & 43.2 & \textbf{39.8} & 40.2 & 45.1 & 43.1 & 43.6 & 50.5 & 50.9 & 44.5 \\
\bottomrule
\end{tabular}
\vspace{-1.0cm}
\end{center}
\end{table}

\setlength{\tabcolsep}{6.5pt}
\begin{table}[h]
\begin{center}
\caption{Effect of which layer to insert masks. ``1$\sim$4'' means layers 1 to 4 are inserted a mask. ``9$\sim$12'' achieves highest performance. The attention in the later layers have a significant impact on shaping the output embedding. The best results are in \textbf{bold}.}
\label{table:layer}
\begin{tabular}{ccccccccccc}
\toprule
\multirow{2}{*}{Layer} & \multicolumn{3}{c}{RefCOCO} & \multicolumn{3}{c}{RefCOCO+} & \multicolumn{2}{c}{RefCOCOg} & \multirow{2}{*}{Avg}\\ 
& TestA & TestB & Val & TestA & TestB & Val & Test & Val \\
\midrule
1 & 17.1 & 25.8 & 20.6 & 17.3 & 26.8 & 20.6 & 24.6 & 26.8 & 22.4 \\
1$\sim$4 & 20.4 & 26.1 & 21.0 & 21.0 & 27.1 & 21.7 & 27.6 & 27.3 & 24.0\\
1$\sim$6 & 22.3 & 25.1 & 22.4 & 22.1 & 25.7 & 23.6 & 28.6 & 28.2 & 24.7 \\
12 & 39.4 & \textbf{40.0} & 39.7 & 43.7 & \textbf{43.8} & 42.9 & 50.9 & 50.6 & 43.9 \\
9$\sim$12 & \textbf{44.2} & 39.4 & 40.8 & \textbf{46.8} & 43.1 & \textbf{44.5} & \textbf{51.5} & \textbf{51.3} & \textbf{45.2} \\
7$\sim$12 & 43.8 & 39.4 & \textbf{41.3} & 46.3 & 42.5 & 44.2 & 51.0 & 51.1 & 44.9 \\
\bottomrule
\end{tabular}
\end{center}
\end{table}

\subsubsection{Image Classification.} 
\begin{figure}[t]
\centering
\includegraphics[scale=0.65]{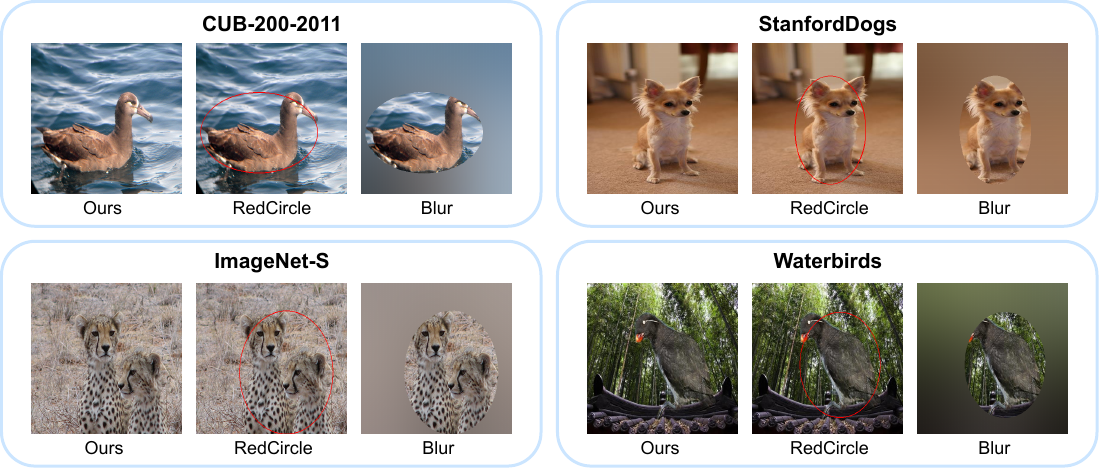}
\caption{Examples of input images in each dataset. For each dataset, from the left to right is the input image of model for our method, RedCircle and Blur respectively.
}
\label{fig:a7}
\vspace{0.2cm}
\end{figure}

\begin{figure}[t]
\centering
\includegraphics[scale=0.7]{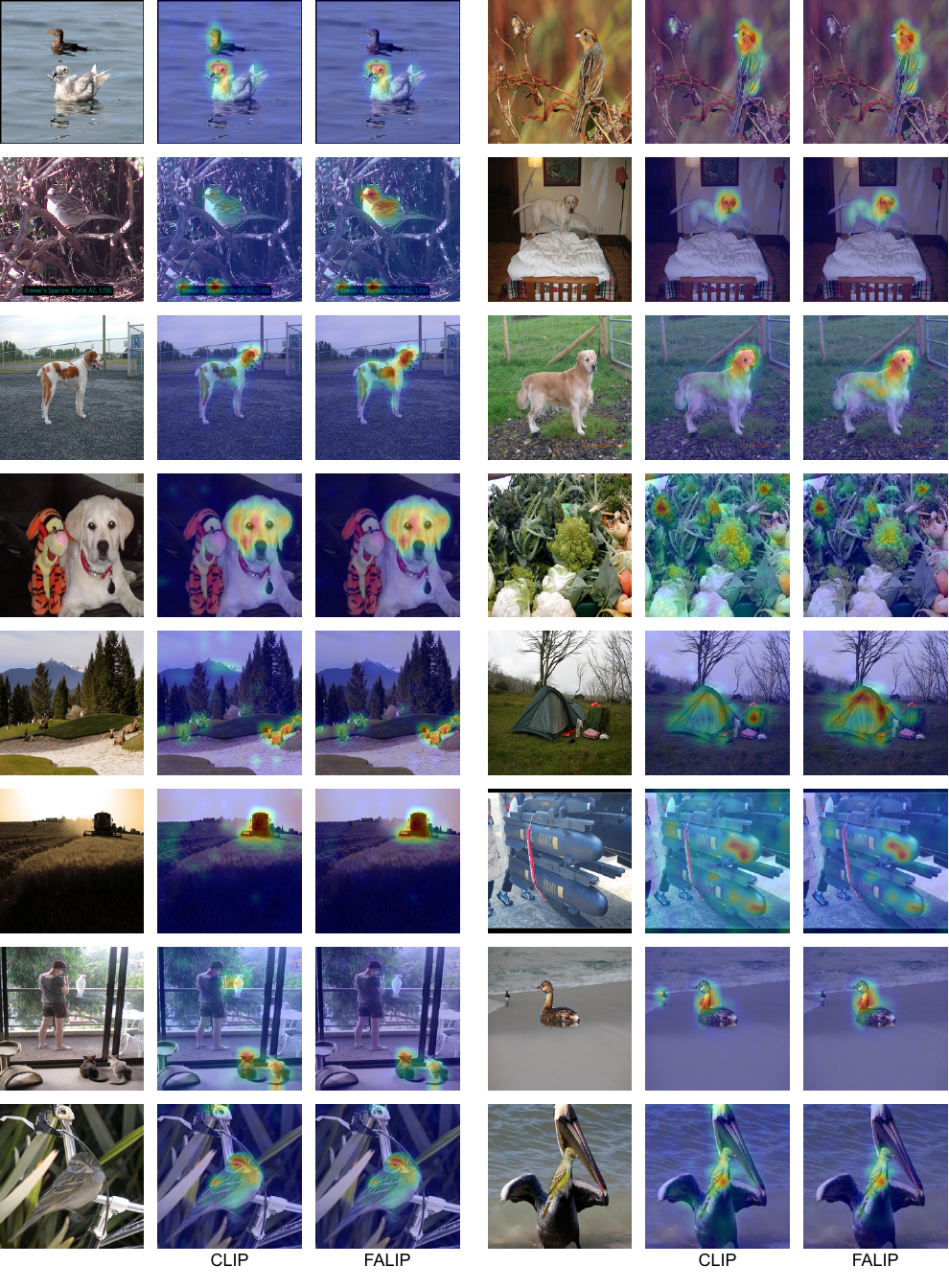}
\caption{Attention visualization. Our model demonstrates its ability to better focus on the target objects rather than irrelevant objects in the background.
}
\label{fig:a1}
\end{figure}

The image classification experimental results are obtained from testing on the following datasets:
entire StanfordDogs, entire CUB-200-2011, test of Waterbirds and validation of ImageNets, which are shown in \cref{table:data_class}. \cref{fig:a7} shows the input image of various methods. \cref{table:mclass} demonstrates the performance of FALIP on the larger model Vit-L/14, showing an improvement over CLIP in terms of accuracy. Except for the Waterbirds, FALIP achieves the highest accuracy on all other datasets. \cref{table:enlargered} illustrates how accuracy is affected by visual prompt of varying sizes. Increasing the range of the RedCircle appropriately can lead to a certain improvement in accuracy. \cref{fig:a2} provides a brief explanation of enlarging size of visual prompt (the maximum size will not exceed the inscribed circle of the image). In \cref{fig:a1} we compare our method with CLIP on the model's attention.

\setlength{\tabcolsep}{4pt}
\begin{table}[t]
\begin{center}
\caption{Method ablation on Image Classification. The best results are in \textbf{bold}, and sub-optimal results are \underline{underlined}.}
\label{table:mclass}
\begin{tabular}{lcccccccccc}

\toprule
\multirow{2}{*}{Method} & \multirow{2}{*}{Model} & \multicolumn{2}{c}{StanfordDogs} & \multicolumn{2}{c}{CUB-200-2011} & \multicolumn{2}{c}{ImageNet-S} & \ {Waterbirds} \\
& & Top1 & Top5 & Top1 & Top5 & Top1 & Top5 & Top1 \\
\midrule
Original CLIP & ViT-B & \underline{56.5}& \underline{85.2} & \underline{54.2} & \textbf{83.7} & \underline{64.9} & \underline{88.4} & \underline{78.2} \\
RedCircle & ViT-B & 52.4 & 82.8 & 44.2 & 77.0 & 62.8 & 86.5 & 77.5\\
Blur & ViT-B & 51.9 & 81.9 & 39.1 & 71.0 &53.8 &77.6 & 78.1\\
\rowcolor{gray!25}
FALIP(Ours) & ViT-B & \textbf{58.3}& \textbf{86.0} & \textbf{54.3} & \underline{83.6}  & \textbf{67.3} &\textbf{89.9} & \textbf{79.7}\\
\midrule
Original CLIP & ViT-L & \underline{65.4}& \underline{89.1} & \underline{61.4} & \underline{90.1} & \underline{72.0} & \underline{91.1} & 83.3 \\
RedCircle & ViT-L & 63.7 & 88.6 & 56.1 & 87.5 & 70.9 & 90.6 & 80.7\\
Blur & ViT-L & 60.1 & 85.4 & 46.1 & 82.8 & 63.6 & 84.2 & \textbf{85.1}\\
\rowcolor{gray!25}
FALIP(Ours) & ViT-L & \textbf{66.6}& \textbf{89.8} & \textbf{61.7} & \textbf{90.7}  & \textbf{74.8} &\textbf{92.7} & \underline{84.5}\\

\bottomrule
\end{tabular}
\end{center}
\end{table}

\begin{figure}[t]
\centering
\includegraphics[scale=0.65]{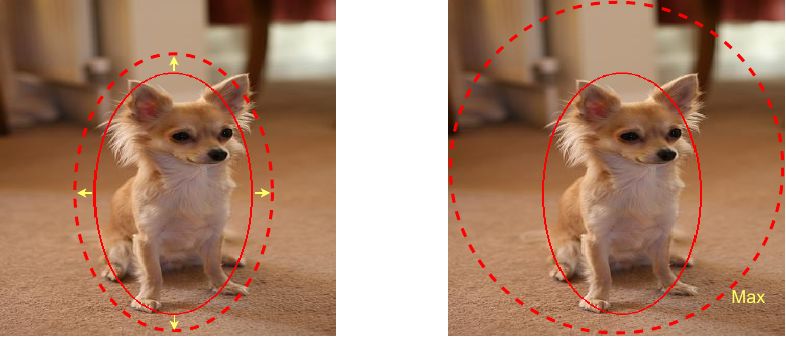}
\caption{Enlarge prompts. We increase the pixels in four directions. In this way, the contamination of foreground can be mitigated. 
}
\label{fig:a2}
\vspace{0.2cm}
\end{figure}
\clearpage

\setlength{\tabcolsep}{7pt}
\begin{table}[t]
\begin{center}
\caption{Method ablation on size of RedCircle. The best results are in \textbf{bold}.}
\label{table:enlargered}
\begin{tabular}{ccccccccccc}

\toprule
\multirow{2}{*}{Enlarge Pixels}  & \multicolumn{2}{c}{StanfordDogs} & \multicolumn{2}{c}{CUB-200-2011} & \multicolumn{2}{c}{ImageNet-S} & \ {Waterbirds} \\
& Top1 & Top5 & Top1 & Top5 & Top1 & Top5 & Top1 \\
\midrule
0  & 52.4 & 82.8 & 44.2 & 77.0 & 62.8 & 86.5 & 77.5\\
5  & 51.8 & 81.8 & 43.2 & 76.0 & 63.2 & 87.2 & 77.6\\
10 & 52.4 & 82.1 & 43.8 & 76.4 & 63.6 & 87.3 & 77.7\\
20 & 52.7 & 82.4 & 45.6 & 77.3 & \textbf{64.3} & 87.7 & 78.0\\
30 & 53.1 & 82.4 & 46.5 & 78.0 & 64.2 & \textbf{88.1} & 78.4 \\
40 & \textbf{53.2} & 82.6 & 47.1 & 78.6 & 64.1 & 87.9 & 78.7\\
50 & 53.0 & \textbf{82.7} & 46.9 & 78.8 & 63.9 & 87.6 & 78.7\\
100 & 52.9 & 82.5 & 47.6 & \textbf{79.0} & 62.6 & 86.7 & 78.6\\
150 & 52.8 & 82.4 & 47.7 & 78.7 & 61.8 & 86.6 & 78.7\\
200 & 52.8 & 82.4 & \textbf{47.8} & 78.9 & 61.7 & 86.2 & \textbf{78.7}\\
\bottomrule
\end{tabular}
\end{center}
\vspace{-1.0cm}
\end{table}

\begin{figure}[t]
\centering
\includegraphics[scale=0.55]{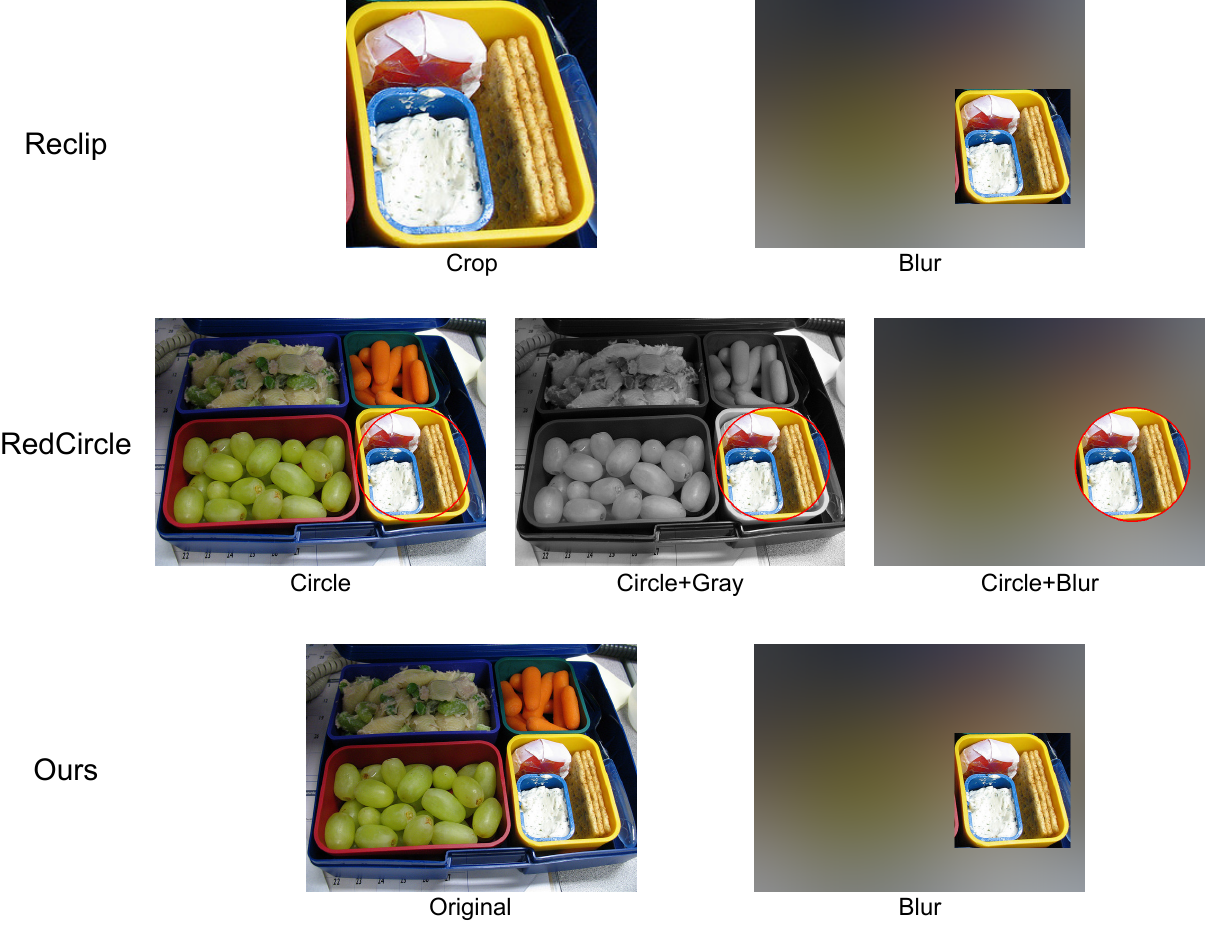}
\caption{The specific approaches for ensemble. To ensure a fair comparison, we also adopt the same Blur method used in the previous method.
}
\label{fig:a2}
\end{figure}
\clearpage

\subsubsection{Pesudo Code.} The pesudo code of FALIP is shown in \textbf{Algorithm 1}.



\begin{algorithm}[h]
\caption{Image Encoder of Foveal-Attention CLIP}
\textbf{Input}: image $x$, bounding box $box$\\  
\textbf{Output}: image feature $f_v$
\begin{algorithmic}[1]

\Function{FALIP}{$x$, $box$}
\State $x^* \leftarrow \text{Preprocess}(x)$
\State $X \leftarrow \text{PatchEmbedding}(x^*) $ \quad\quad \textcolor{blue!30!cyan}{\#Transform image to sequence, $X\in\mathbb{R}^{(N+1)\times D}$}
\State $T \leftarrow \text{BoxToToken}(x, box) $ \quad\quad \textcolor{blue!30!cyan}{\#Transform box to token space}
\State $H, W \leftarrow T.height, T.wdith$ 
\State $R \leftarrow \vmathbb{0}^{H\times W}$  \quad\quad \textcolor{blue!30!cyan}{\#Initialize with 0 }
\State $M \leftarrow \vmathbb{0}^{(N+1)\times (N+1)}$ \quad\quad \textcolor{blue!30!cyan}{\#Initialize with 0, $N+1$ is length of the sequence}
\For{$i = 0$ to $(H-1)$}
\For{$j = 0$ to $(W-1)$}
  \State $R[i][j] \leftarrow e^{-\frac{[i-(H-1)/2]^2+[j-(W-1)/2]^2}{2\sigma^2}}$ \quad\quad \textcolor{blue!30!cyan}{\#Generate foveal value}
  \EndFor 
\EndFor
\State $R^{norm} \leftarrow \alpha \times \frac{R-\text{Min}(R)+\epsilon}{\text{Max}(R)-\text{Min}(R)+\epsilon}$ \quad\quad \textcolor{blue!30!cyan}{\#Normalization}
\State $R^* \leftarrow \text{Flatten}(R^{norm})$ \quad\quad \textcolor{blue!30!cyan}{\#Flatten and align indices with $X$}
\State $M[0] \leftarrow R^*$ \quad\quad \textcolor{blue!30!cyan}{\#Assgin value to positions in the first row of $M$ }
\State $X^* \leftarrow \text{LayerNorm}(X)$ 
\State $f_v \leftarrow \text{Transformer}(X^*,M)$ \quad\quad 
\textcolor{blue!30!cyan}{\#Input sequence and foveal attention mask}

\EndFunction
\end{algorithmic}
\end{algorithm}


\end{document}